\documentclass[10pt,journal,compsoc]{IEEEtran}
\usepackage{cite}
\usepackage[pdftex]{graphicx}
\usepackage{amsmath}
\usepackage{algorithmic}
\usepackage{array}
\usepackage[caption=false,font=footnotesize,labelfont=sf,textfont=sf]{subfig}
\usepackage{url}
\usepackage[backref=section,
colorlinks=true,
citecolor=blue,
linkcolor=red,
anchorcolor=red,
unicode=true,
urlcolor=blue]{hyperref}
\usepackage{soul}
\usepackage{color}
\usepackage{parskip}
\usepackage{bm}
\usepackage{bbm}
\usepackage{mathtools}
\usepackage{amsthm}
\usepackage{amssymb}
\usepackage{amsfonts}
\usepackage{babel}
\usepackage{booktabs}
\usepackage{multirow}
\usepackage{enumitem}
\usepackage{pdfpages}
\usepackage{algorithm}

\makeatletter
\theoremstyle{remark}
\newtheorem*{rem*}{\protect\remarkname}
\theoremstyle{plain}

\theoremstyle{plain}

\theoremstyle{plain}

\theoremstyle{plain}

\theoremstyle{definition}

\theoremstyle{plain}

\theoremstyle{definition}

\theoremstyle{definition}

\providecommand{\assumptionname}{Assumption}
\providecommand{\corollaryname}{Corollary}
\providecommand{\definitionname}{Definition}
\providecommand{\lemmaname}{Lemma}
\providecommand{\propositionname}{Proposition}
\providecommand{\remarkname}{Remark}
\providecommand{\theoremname}{Theorem}
\providecommand{\conditionname}{Condition}
\providecommand{\examplename}{Example}

\makeatletter
\newcommand{\removelatexerror}{\let\@latex@error\@gobble}

\begin{document}
\title{Entity-Level Text-Guided Image Manipulation}
\author{
Yikai~Wang$^{*}$,
Jianan~Wang$^{*}$, 
Guansong~Lu,
Hang~Xu,
Zhenguo~Li,
Wei~Zhang, 
and Yanwei~Fu.
\IEEEcompsocitemizethanks{
\IEEEcompsocthanksitem Yikai Wang and Jianan Wang contribute equally.
\IEEEcompsocthanksitem Yikai Wang, Jianan Wang, and Yanwei Fu are with the School of Data Science, Fudan University. E-mail: \{yikaiwang19, jawang19, yanweifu\}@fudan.edu.cn
\IEEEcompsocthanksitem Guansong Lu, Hang Xu, Zhenguo Li, and Wei Zhang are with Huawei Noah’s Ark Lab. E-mail: \{luguansong, xu.hang, li.zhenguo, wz.zhang\}@huawei.com
}
}

\markboth{Journal of \LaTeX\ Class Files,~Vol.~14, No.~8, August~2015}%
{Shell \MakeLowercase{\textit{et al.}}: Bare Demo of IEEEtran.cls for Computer Society Journals}

\IEEEtitleabstractindextext{%
\begin{abstract}
Existing text-guided image manipulation methods aim to modify the appearance of the image or to edit a few objects in a virtual or simple scenario, which is far from practical applications. 
In this work, we study a novel task on text-guided image manipulation on the entity level in the real world (eL-TGIM). 
The task imposes three basic requirements, 
(1) to edit the entity consistent with the text descriptions, 
(2) to preserve the entity-irrelevant regions, and 
(3) to merge the manipulated entity into the image naturally. 
To this end, we propose an elegant framework, dubbed as \emph{SeMani}, forming the \emph{Se}mantic \emph{Mani}pulation of real-world images that can not only edit the appearance of entities but also generate new entities corresponding to the text guidance. 
To solve eL-TGIM, SeMani decomposes the task into two phases:
the semantic alignment phase and the image manipulation phase.
In the semantic alignment phase, SeMani incorporates a semantic alignment module to locate the entity-relevant region to be manipulated.
In the image manipulation phase, SeMani adopts a generative model to synthesize new images conditioned on the entity-irrelevant regions and target text descriptions.
We discuss and propose two popular generation processes that can be utilized in SeMani, the discrete auto-regressive generation with transformers and the continuous denoising generation with diffusion models, yielding SeMani-Trans and SeMani-Diff, respectively.
We conduct extensive experiments on the real datasets CUB, Oxford, and COCO datasets to verify that SeMani can distinguish the entity-relevant and -irrelevant regions and achieve more precise and flexible manipulation in a zero-shot manner compared with baseline methods.
Our codes and models will be released at \url{https://github.com/Yikai-Wang/SeMani}.
\end{abstract}
}
\maketitle
\IEEEdisplaynontitleabstractindextext
\IEEEpeerreviewmaketitle

\section{Introduction}
\label{sec:intro}
\begin{figure}
\centering
\includegraphics[width=\linewidth]{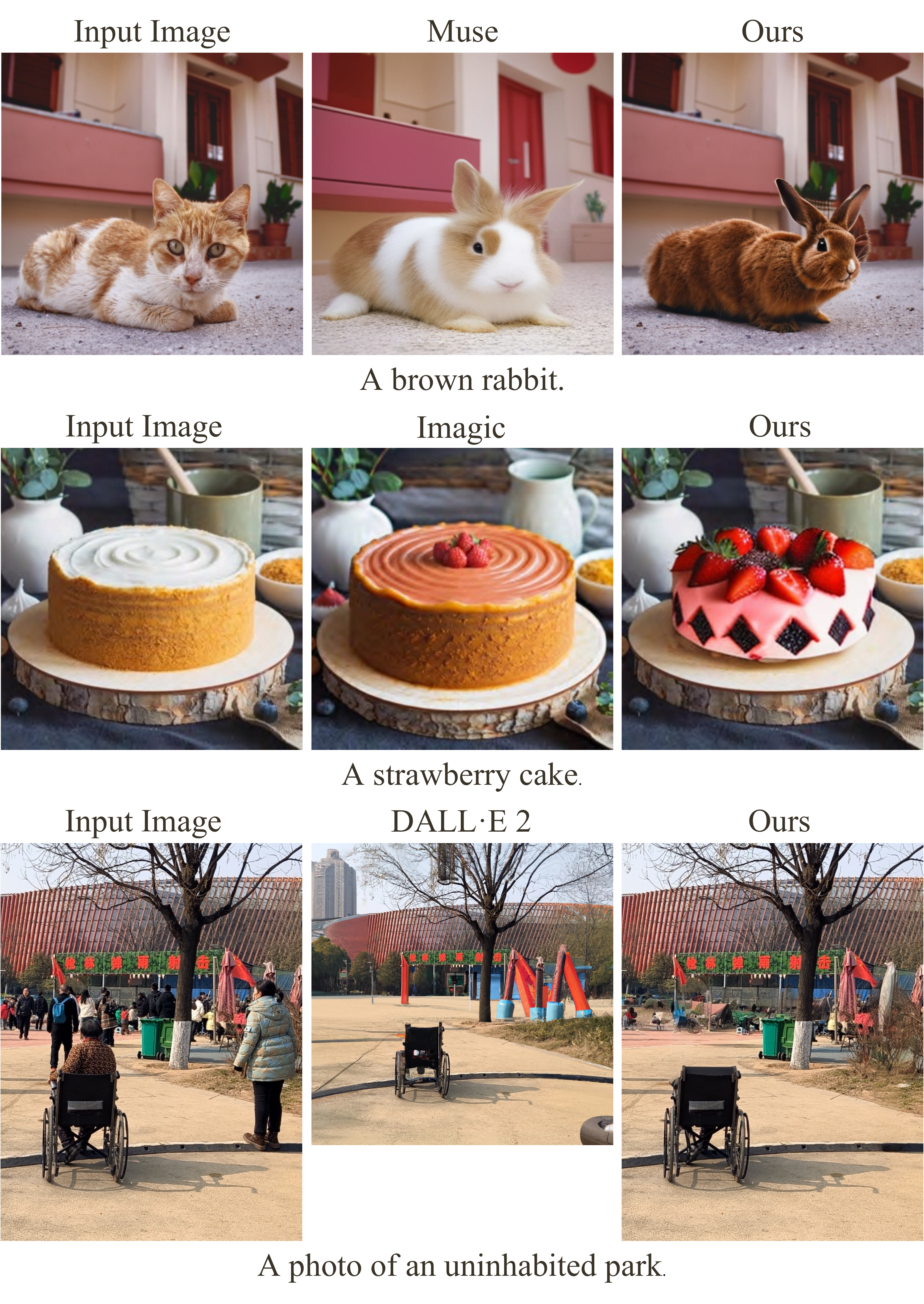}
\vspace{-1cm}
\caption{
Comparison between our method and powerful image editors, including transformer-based architectures (Muse~\cite{chang2023muse}), diffusion-based pixel-level model (Imagen~\cite{sahariaphotorealistic} based Imagic~\cite{kawar2022imagic}) and latent-level model (DALLE2~\cite{ramesh2022hierarchical}).
Analyses are in the second paragraph of Sec.~\ref{sec:intro}.
}
\label{fig:teaser-diffusion}
\end{figure}

\begin{figure}
\centering
\includegraphics[width=\linewidth]{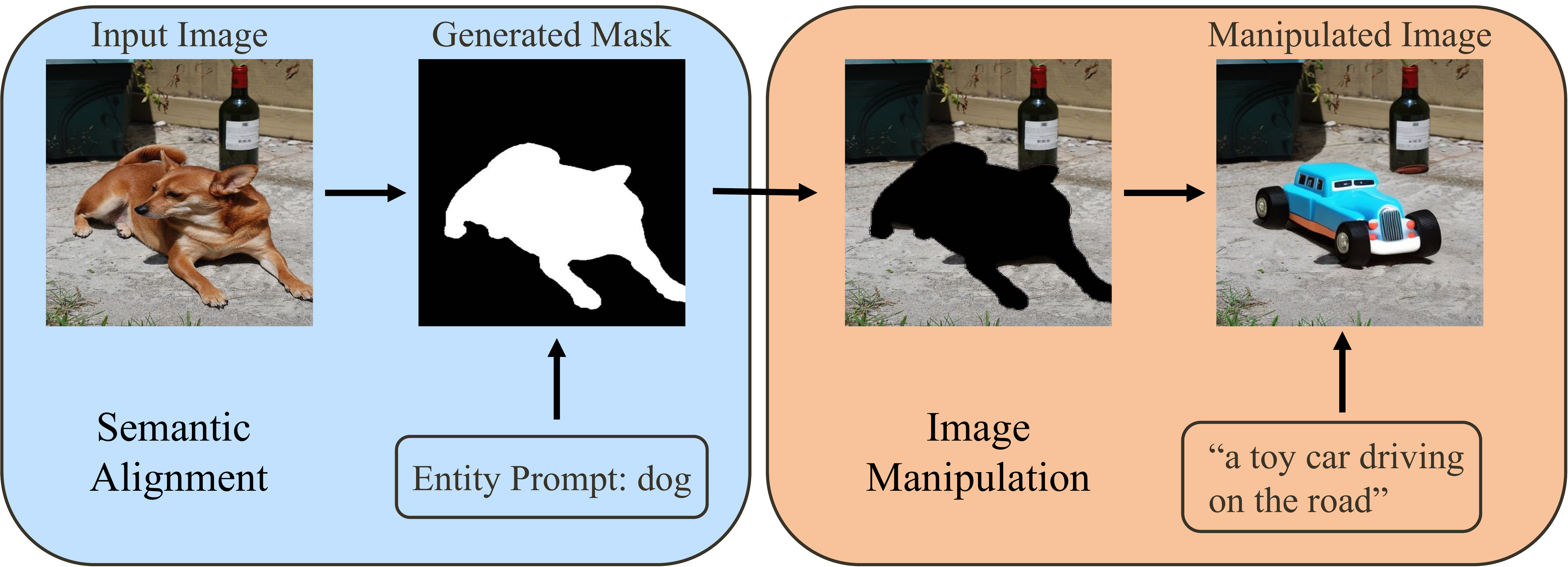}
\caption{Pipeline of SeMani for entity-Level Text-Guided Image Manipulation, which is decomposed into the semantic alignment phase and image manipulation phase.
In the semantic alignment phase, we focus on the entity-relevant region of the input image given the entity prompt and generate the entity mask to locate the entity in the image.
In the image manipulation phase, we generate new images via the target description while preserving the entity-irrelevant regions of the image.
}
\label{fig:eL-TGIM}
\end{figure}

\IEEEPARstart{T}{here} are various active branches of image manipulation, such as style transfer \cite{style_transfer}, image translation \cite{isola2017pix2pix,zhu2017cyclegan}, 
and Text-Guided Image Manipulation (TGIM),
by taking advantage of recent deep generative architectures such as GANs~\cite{2014Generative}, VAEs~\cite{2014Auto}, auto-regressive models~\cite{vaswani2017attention} and diffusion models~\cite{sohl2015deep}. 
Particularly, the previous TGIM methods either operate some objects by  text instructions \cite{el2019tell,zhang2021text,fu2020iterative}, such as ``adding'' and ``removing'' in a simple toy scene, or manipulating the appearance 
of objects\cite{chen2018language} or the style of the image\cite{wang2018learning,jiang2021language}.
In this work, we are interested in a novel challenging  task of 
entity-Level Text-Guided Image Manipulation (eL-TGIM), which is to manipulate the entities on a natural image given the text descriptions,
as shown in Fig.~\ref{fig:eL-TGIM}.  
eL-TGIM takes as inputs the image to be manipulated, a word prompt to locate the interested entity, and a target text description to manipulate the entity.
Basically, eL-TGIM imposes three requirements:
(1) to edit the entity consistent with the text
descriptions, (2) to preserve the entity-irrelevant regions, and (3) to merge the manipulated entity into the image naturally. 
Critically, 
our eL-TGIM is much more difficult than the vanilla TGIM task, as it demands manipulation ability at the fine-grained entity level. Thus, it is nontrivial to directly extend previous methods to the eL-TGIM task, as they can not effectively identify and edit the properties of entities.

\begin{figure*}
\centering
\includegraphics[width=\linewidth]{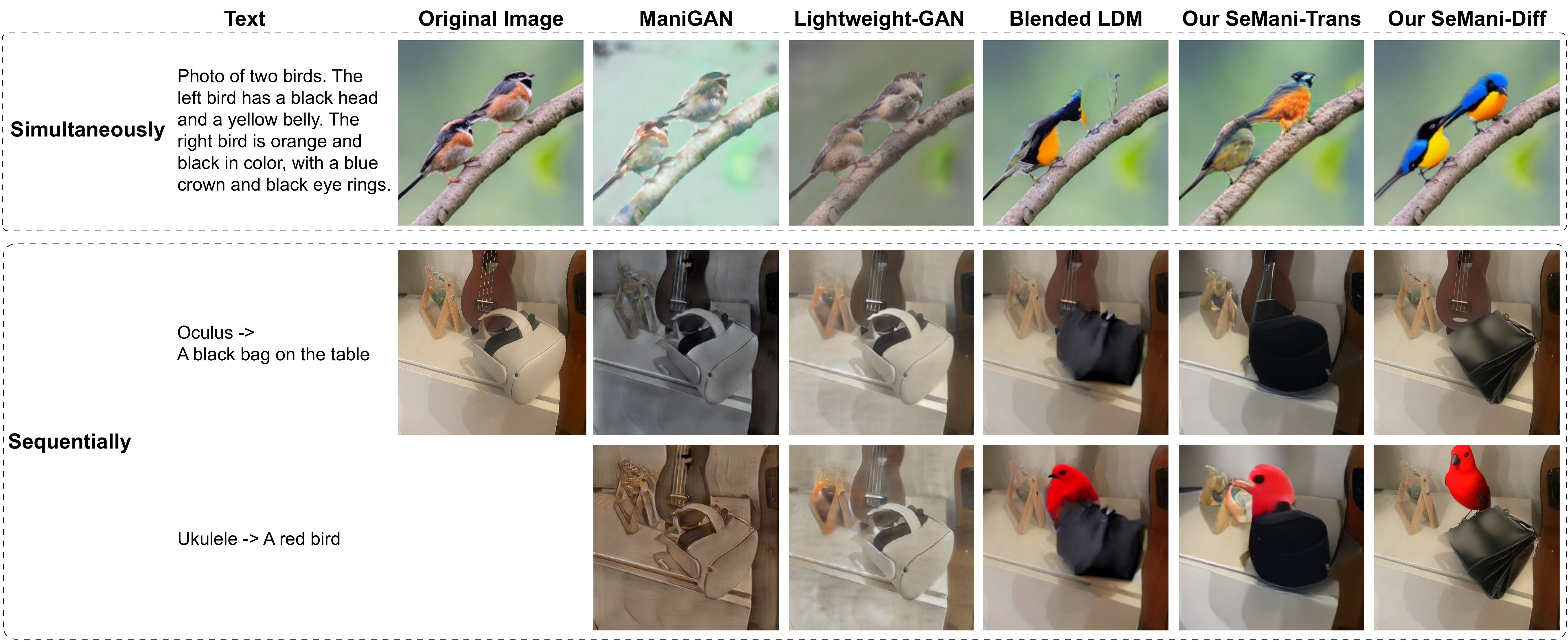}
\caption{Results of manipulating multiple objects simultaneously or sequentially. 
Blended LDM uses the masks generated by SeMani as it requires user-provided masks.
SeMani can manipulate different objects consistently with different texts, while competitors cannot.
}
\label{fig:multiple_object}
\end{figure*}

To demonstrate the difficulty of eL-TGIM, we adopt several recent powerful image editors, including transformer-based architectures (Muse~\cite{chang2023muse}), diffusion-based pixel-level model (Imagen~\cite{sahariaphotorealistic} based Imagic~\cite{kawar2022imagic}) and latent-level model (DALLE2~\cite{ramesh2022hierarchical}).
Results are shown in Fig.~\ref{fig:teaser-diffusion}, 
where the left column is the input image, the middle column is the manipulation results by other methods, and the right column is our manipulation results.
The corresponding text description is at the bottom.
Results of Muse and Imagic are reported in their paper, and results of DALLE2 are generated using the official API.
Powerful editing algorithms can only satisfy partial requirements of eL-TGIM, which can generate new images that are consistent with text description with the cost of neglecting some essential components for eL-TGIM.
For example, 
Muse and Imagic will change the background (see the door of the first row and the cup of the second row), and DALLE2 requires a user-provided mask to detect the regions to be manipulated and can only generate square images.
Our method consistently satisfy the three requirements of eL-TGIM with only text guidance.

Generally, the major obstacle of the TGIM task lies in distinguishing which parts of the image to change or not change. 
To tackle this problem, existing TGIM methods  \cite{sisgan,tagan,li2020manigan,li2020lightweight} 
propose many different manipulation mechanisms, such as word-level discriminator \cite{tagan,li2020lightweight} and text-image affine combination module \cite{li2020manigan}, to differentiate the candidate editing regions from the other image parts.
These methods unfortunately are still very limited to be applied to manipulate the entities in nature images. For example,
Fig.~\ref{fig:multiple_object} shows that previous methods can only manipulate the texture/color of an object or require user-provided masks to locate the entity-relevant region, while they fail to perform reasonable entity-level manipulation from text descriptions. 

To this end, we propose a novel framework of Semantic Manipulation (SeMani), which decomposes the eL-TGIM task into the semantic alignment phase and image manipulation phase.
Imitating the human activities of image editing, we first identify the entity-relevant region corresponding to the entity prompt.
The entities should be open vocabulary and not limited to a fixed set of categories.
Our target in this phase is to generate the mask of the entity.
With this mask, we can focus on the entity and preserve entity-irrelevant regions when manipulating the image.
In the second phase, we perform image manipulation with powerful generative models.
The target of this phase is to generate new images that are consistent with the target text description and entity-irrelevant regions.

To implement SeMani, we propose two variants that view the image from different perspectives.
Specifically, our SeMani-Trans view the image as a discrete token sequence and propose a token-wise semantic alignment module to locate the entity-relevant tokens and perform manipulation on the token sequence in an auto-regressive manner. 
On the other hand, our SeMani-Diff focuses on the continuous pixel space and directly provides a pixel-level semantic alignment with a denoising generation with diffusion models to perform image manipulation.

\textbf{SeMani-Trans}. 
Recently, transformer-based
models~\cite{van2017neural,vqvae2,vqgan} have been proposed for image synthesis and have shown great expressive power. 
We thus present a transformer model for generation, by first learning an auto-encoder to down-sample and quantize an image as a sequence of discrete image tokens and then fit the joint distribution of this sequence with a transformer-based auto-regressive model.
Furthermore, to successfully identify the entities for editing, our semantic alignment modules include a patch-level segmentation model and a Contrastive Language-Image Pretraining (CLIP~\cite{radford2021clip}) model. 
The former model helps to
locate the entities that existed in the image and the latter model helps to identify the text-relevant image tokens given the textual guidance.  
Thus our SeMani-Trans generation model can manipulate the image locally and preserve the irrelevant contents to a greater extent, as in Fig.~\ref{fig:multiple_object}. 
On the other hand, we repurpose the CLIP model as one type of semantic loss to further boost the visual-semantic alignment between the input textual guidance and the manipulated image. 
Essentially, such semantic loss, proposed in our SeMani-Trans, is complementary to token-wise classification loss, and thus efficiently serves as a pixel-level supervision signal to train our model. 

\textbf{SeMani-Diff} provides a continuous alternative to SeMani-Trans and solves several issues that existed in SeMani-Trans.

\textbf{(1)} The auto-regressive generation pipeline limits the available knowledge for the generation model to utilize in the eL-TGIM task.
In SeMani-Trans, we encode the image to a token sequence, where the prediction is performed step by step along the sequence.
When the entity-relevant region is in the latter part of the sequence, SeMani-Trans could utilize most of the knowledge in the sequence to provide a more precise and consistent manipulation.
However, when the entity-relevant region lies in the early part of the sequence, or extremely at the start of the sequence, SeMani-Trans could only utilize very little knowledge to help generation.
This inherent drawback of auto-regressive generation limits the generation capacity of SeMani-Trans.
To solve this issue, we adopt the recently fast-developed denoising diffusion probabilistic models (DDPMs)~\cite{sohl2015deep} to perform generation given the knowledge of the whole image instead of a uni-direction.
DDPMs construct the connection between random noises and real images, performing generation via the denoising process.
In each step, the knowledge of the whole image can be utilized by DDPMs to perform generation.

\textbf{(2)} The locally calculated similarity is sub-optimal for the semantic alignment module.
In SeMani-Trans, the similarity between the visual features of entities and semantics is calculated via an averaged visual token similarity with the semantic token.
However, the visual feature of each token considers more local patterns instead of global patterns, and a simple average may not well-extract the relation between tokens of the entity.
On the other hand, the global feature for the entire entity is preferable, but the shape of the entity has various types. 
However, CLIP-like models are trained on images with square shapes.
To this end, we strike the balance between local similarity and global similarity and propose to fine-tune the CLIP model with entity image, where entity-irrelevant regions are masked to ensure that only patterns of the entity are extracted.
With this modification, the visual features of the entity will include more global and thus semantic information, which benefits the alignment with words.

With these modifications, our SeMani-Diff first adopts the segmentation model to locate several entities in the image, then the entities will be encoded via the fine-tuned CLIP model.
The semantic alignment module will select the most-possible entity based on the consistency with the entity prompt in a more global way.
Then, SeMani-Diff uses DDPMs to manipulate the entity-relevant region with entity-irrelevant regions and target text descriptions.

We evaluate both SeMani-Trans and SeMani-Diff on multiple datasets including CUB~\cite{cub200}, Oxford~\cite{oxford102}, and COCO~\cite{mscoco}. 
Quantitatively, qualitatively, and user-study comparisons against previous methods demonstrate the superiority of SeMani in all three requirements of eL-TGIM. 

\textbf{Contributions}. 
In summary, our contributions are:
\begin{itemize}[leftmargin=*,itemsep=0pt,topsep=0pt,parsep=0pt]
\item We introduce a new task, entity-Level Text-Guided Image Manipulation (eL-TGIM) which aims to manipulate entities of the image with only text descriptions.
\item To solve eL-TGIM, we propose an elegant SeMani framework, that decomposes the eL-TGIM into the semantic alignment phase and image manipulation phase. 
\item We propose a 
transformer-based framework with discrete token-wise semantic alignment and generation, named SeMani-Trans, which can not only manipulate the texture/color of a single object but also manipulate the structure of an object and manipulate multiple objects. 
\item We further improve the generation process and semantic alignment module by proposing SeMani-Diff, which runs prediction with knowledge of the whole images instead of a uni-direction direction and extracts visual features of entities in a more global way.
\item We quantitatively and qualitatively evaluate our method on the CUB, Oxford, and COCO datasets, achieving better results against baseline methods.
\end{itemize}
\textbf{Extension}. Our conference version of this work was published in~\cite{wang2022manitrans} as an oral paper. 
Compared with~\cite{wang2022manitrans}, we have the following extensions.
\begin{itemize}[leftmargin=*,itemsep=0pt,topsep=0pt,parsep=0pt]
\item We generalize the ideology of~\cite{wang2022manitrans} and show that the decomposition of semantic alignment and image manipulation is essential for eL-TGIM.
\item We analyze several limitations of models in~\cite{wang2022manitrans}, and propose the corresponding improvements for better manipulation capacity.
\item Based on the improvements, we propose a new SeMani-Diff framework with a more global semantic alignment module and a better generation pipeline to utilize knowledge of unmasked regions.
\item Our proposed SeMani-Diff shows the superior quantitative, qualitative, and human evaluation performance of the eL-TGIM task on CUB, Oxford, and COCO datasets. 
\end{itemize}
\section{Related work}

\textbf{Text-to-image generation}
focuses on generating images to visualize what texts describe. There are many good GAN-based models  \cite{GAN-INT-CLS, xu2018attngan,zhang2017stackgan,zhang2018stackgan++}.
Li \emph{et al.} \cite{li2019controllable} further introduce a word-level discriminator network to provide the generator network with fine-grained feedback. Besides GANs, some works also explore applying transformer-based networks for text-to-image generation \cite{dalle,ding2021cogview,esser2021imagebart}. 
Recently, denoising diffusion probabilistic models (DDPMs)~\cite{sohl2015deep} is fast-developed for text-to-image generation tasks.
DDPMs bridge the connection between the distribution of real images and the random Gaussian distribution, such that one can start with random Gaussian noise and iteratively denoise it to generate the image.
Many variants of DDPMs, including GLIDE~\cite{nichol2022glide}, Cascaded Diffusion~\cite{ho2022cascaded}, Imagen~\cite{saharia2022photorealistic}, DALLE2~\cite{ramesh2022hierarchical},LDM~\cite{rombach2022high}, to name a few, have achieved state-of-the-art performances on text-to-image generation task.
In contrast, rather than generating images according to texts, we focus on entity-level manipulating images given texts.

\textbf{Text-guided image manipulation}
has attracted extensive attention as it enables the users to flexibly edit an image with natural language~\cite{sisgan,tagan,
li2020manigan,li2020lightweight,xia2021tedigan,stylegan,el2019tell,zhang2021text,fu2020iterative,chen2018language,wang2018learning,jiang2021language}
. 
Particularly,
Li~\emph{et al.}~\cite{li2020manigan} introduces a multi-stage network with a novel text-image combination module to generate high-quality images. Li~\emph{et al.}~\cite{li2020lightweight} propose a new word-level discriminator along with explicit word-level supervisory labels to provide the generator with detailed training feedback related to each word, achieving a lightweight and efficient generator network.
Recently, due to the good synthesizing capability of StyleGAN, researchers devote to image manipulation by pre-trained StyleGAN models~\cite{xia2021tedigan,patashnik2021styleclip}. Patashnik \emph{et al.} \cite{patashnik2021styleclip} adopt the CLIP model for semantic alignment between text and image, and propose mapping the text prompts to input-agnostic directions in StyleGAN’s style space, achieving interactive text-driven image manipulation.
Text2LIVE~\cite{bar2022text2live} introduces an edit layer to composite the generation results with the image to preserve the information.
The edit layer is directly predicted by a U-Net model.

Diffusion models also show promising performance in text-guided image manipulation. 
Blended Diffusion~\cite{avrahami2022blended} adopts the user-provided mask and target text description for manipulation.
The target text is utilized via the gradient of CLIP loss to the diffusion outputs to guide the generation of diffusion models.
Some works rely on the textual inversion technique~\cite{gal2022image} to represent the image via the text embedding and then achieve manipulation by an interpolation between target descriptions and inverse text embedding~\cite{kawar2022imagic}.
Others~\cite{meng2022sdedit} rely on the inversion of the image that learns a noise that can be transformed into the input image via diffusion models, and then achieve manipulation with this noise.
Prompt-to-Prompt~\cite{hertz2022prompt} controls the cross-attention layer of diffusion models and injects it with weights that correspond to target text to achieve manipulation.
Prompt-to-Prompt relies on the inversion of an image or improved textual inversion~\cite{mokady2022null} to implement manipulation on real images.
On the contrary, our SeMani can directly manipulate real images.

\textbf{Semantic image synthesis}
aims to generate a photo-realistic image from a semantic label. 
Isola \emph{et al.} \cite{isola2017pix2pix} propose a unified framework based on conditional GANs \cite{mirza2014conditional} for various image-to-image translation tasks, including $Semantic ~ labels \leftrightarrow photo$, $Edges \rightarrow Photo$, $Day \rightarrow Night$, and so on. 
Chen and Koltun \cite{chen2017cascaded_refinement} adopt a modified perceptual loss to synthesize high-resolution images to tackle the instability of adversarial training.
Wang \emph{et al.} \cite{wang2018pix2pixhd} propose a novel adversarial loss and a new multi-scale generator and discriminator architectures for generating high-resolution images with fine details and realistic textures. Park \emph{et al.} \cite{park2019SPADE} propose a spatially-adaptive normalization layer to modulate the activation using input semantic layouts and effectively propagate the semantic information throughout the network.
Such works enable users to synthesize images with a finite number of semantic concepts associated with the semantic labels, while our method focuses on manipulating the input images according to the input texts, which is more flexible and with an unlimited number of semantic concepts.

\begin{figure*}
\centering
\includegraphics[width=\linewidth]{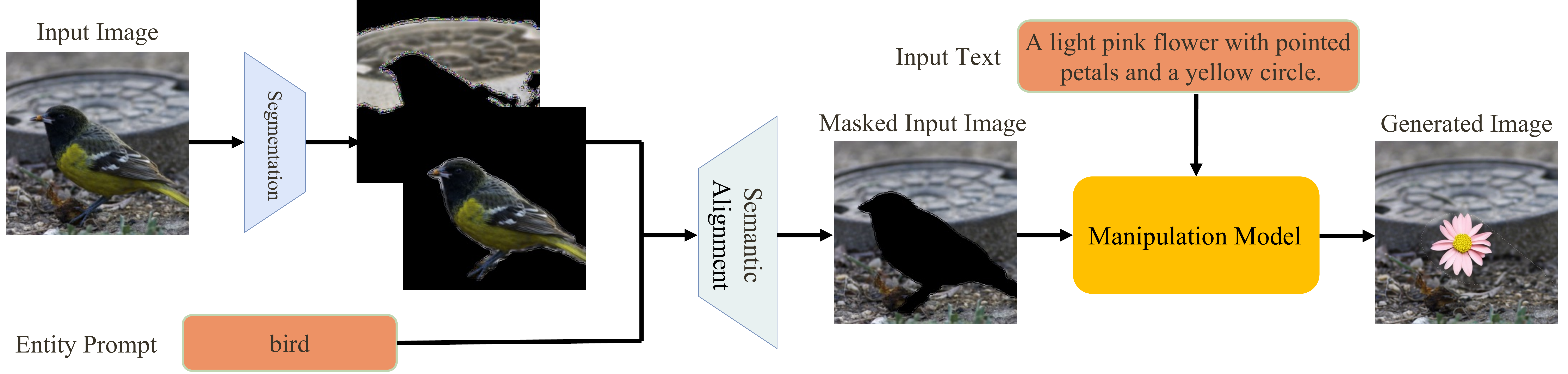}
\caption{Framework of our SeMani. 
We achieve eL-TGIM by first using a segmentation model to distinguish entities that existed in the image, and then adopting a semantic alignment model to identify the prompt-relevant entity.
Then the masked image is provided to condition the manipulation model with target text descriptions to perform semantic manipulation.
}
\label{fig:framework}
\end{figure*}

\textbf{Vision and language representation learning}
 models \cite{radford2021clip,jia2021align,zhang2021vinvl,zhuge2021kaleido,li2020unimo,chen2020uniter,lin2021m6,li2020oscar,su2019vlBERT} learn cross-modal representations for various down-stream tasks, including image-text retrieval, image captioning, visual grounding, and so on. They adopt the network architecture of ResNets \cite{resnet} and/or Transformers \cite{vaswani2017attention,dosovitskiy2020vit}, and mainly use two kinds of learning tasks for pre-training: cross-modal contrastive learning and masked language modeling. Specifically, the recent CLIP \cite{radford2021clip} model is trained on a large-scale dataset and shows superior performance on zero-shot tasks. We repurpose the CLIP model to help train our framework for eL-TGIM.

\section{Methodology}
\subsection{Overview}
\textbf{Problem formulation}.
Entity-Level Text-Guided Image Manipulation (eL-TGIM) aims to perform image editing in the real world, with a focus on the entity level manipulation guided by text.
eL-TGIM imposes three basic requirements:
\begin{enumerate}[leftmargin=*,itemsep=0pt,topsep=0pt,parsep=0pt]
\item To edit the entity consistent with the text
descriptions; 
\item To preserve the entity-irrelevant regions;
\item To merge the manipulated entity into the image naturally. 
\end{enumerate}
Formally, eL-TGIM takes as input the entity prompt word $\bm{e}$, the target text description $\bm{T}$, and the original image $\bm{X}$ from the real world, i.e., $\bm{X}\sim p(\bm{X})$ where $p(\bm{X})$ indicates the distribution of real-world images, 
The target is to generate a new image $\tilde{\bm{X}}$ that follows the above three requirements.

\textbf{SeMani}.
In this paper, we propose SeMani, forming the Semantic Manipulation for eL-TGIM.
Inspired by the human activities for image editing, we decompose the eL-TGIM into the semantic alignment and image manipulation phases.

In the semantic alignment phase, we aim to generate a mask $\bm{M}\in\{0,1\}^{\bm{X}}$ for the original image $\bm{X}$, such that the resulting $\bm{M}\odot\bm{X}$ only contains the interested entity and all other regions are masked, where $\odot$ indicates element-wise multiplication.
To achieve this, SeMani first adopts a segmentation model to generate a series of masks $\{\bm{M}_i\}$ such that each mask indicates an entity existed in the image $\bm{X}$.
Then, given the entity prompt $\bm{e}$, we propose a semantic alignment module to identify the most possible entity via
\begin{equation}
\bm{M}_e\coloneqq \mathrm{argmax}_{\bm{M}_i}\  \mathrm{Sim}(\bm{M}_i\odot\bm{X},\bm{e}),
\end{equation}
where the similarity function $\mathrm{Sim}(\cdot)$ is defined with a modified CLIP model specifically for architectures we used, which will be introduced later.
Note that the entity prompt here should be \emph{open vocabulary} and not limited to a fixed set of categories for better practical utilization.

In the image manipulation phase, SeMani takes as input the masked image $\bm{M}_e\odot\bm{X}$ and the target text description $\bm{T}$ to generate new image $\tilde{\bm{X}}$ with a deep manipulation model.

With this framework, as illustrated in Fig.~\ref{fig:framework}, SeMani achieves eL-TGIM for input images with text guidance.
In SeMani, the three requirements of eL-TGIM become:
\begin{equation}
\begin{aligned}
 \max\ & \mathrm{Sim}(\tilde{\bm{X}},\bm{T})\\
 s.t.\ & \bm{M}_{\bm{e}}\odot\bm{X}\approx \bm{e},\\
 &(1-\bm{M}_{\bm{e}})\odot\tilde{\bm{X}}=(1-\bm{M}_{\bm{e}})\odot\bm{X},\\
 &\tilde{\bm{X}}\sim p(\bm{X}),  
\end{aligned}
\end{equation}
where $\approx$ indicates the consistency between entity prompt and visual object, $=$ indicates pixel-level equivalent, and 
$\tilde{\bm{X}}\sim p(\bm{X})$ indicates $\tilde{\bm{X}}$ follows the distribution of the real image, thus satisfying the third requirement of eL-TGIM.

\textbf{Implementing SeMani}.
To implement SeMani, we resort to the two popular perspectives of viewing images, the discrete and the continuous perspectives.

The discrete perspective of viewing images is inspired by the recent development of natural language processing, mostly BERT~\cite{devlin2018bert} and GPT~\cite{radford2018improving}.
Researchers want to introduce the success of NLP models to the vision field by transforming the image into a visual token sequence.
Then one can treat the image as a visual ``sentence'' and perform NLP techniques on the visual sentence to do vision tasks.
In this paper, we follow this direction and propose a discrete variant of SeMani, SeMani-Trans, to achieve eL-TGIM discretely.
Specifically, we train an auto-encoder to learn a discrete codebook such that any image can be transferred into a discrete token sequence with this codebook vocabulary.
For the semantic alignment module, a token-level CLIP model is required to calculate token similarity.
For image manipulation, an auto-regressive prediction for generation is implemented to achieve generation on the token space.

The continuous perspective of viewing images is the most straightforward way to model images.
Models that follow this perspective directly run on the pixel-level of images and can perform more precise control for semantic alignment and image manipulation.
For the semantic alignment, as the comparison is between masked image $\bm{M}_i\odot\bm{X}$ and entity prompt $\bm{e}$, a fine-tuning of CLIP model is needed to better extract the visual features of images with masks.
For image manipulation, we resort to the recently fast-developed denoising diffusion probabilistic models~\cite{sohl2015deep} (DDPMs) to perform generation by denoising from the pixel-level random noise iteratively.
These modules form our continuous variant of SeMani, SeMani-Diff.

In the following, we introduce the architectures and training details of SeMani-Trans and SeMani-Diff, respectively.

\subsection{SeMani-Trans}
\subsubsection{Architectures}

\textbf{The autoencoder} consists of three components, a convolutional encoder $E$, a convolutional decoder $G$ and a codebook $\bm{Z} \in \mathbb{R}^{K \times n_z}$, 
containing $K$ $n_z$-dimensional latent variables. All of them are learnable.
Given an image $\bm{X} \in \mathbb{R}^{H \times W \times 3}$, $E$ encodes the image into a 
two-dimensional latent feature map $\bm{Q} \in \mathbb{R}^{h \times w \times n_z}$. The codebook is utilized to quantize the latent feature map by replacing each pixel embedding with its closest latent variables within the codebook element-wisely as follows:
\begin{equation}
    \bm{\hat{Q}}_{ij} = \arg\min_{\bm{z}_k}\parallel \bm{Q}_{ij} - \bm{z}_{k} \parallel^{2}.
\end{equation}
For reconstruction, the decoder $G$ takes the quantized latent feature map $\bm{\hat{Q}}$ as input and returns an generated image $\bm{\hat{X}}$ close to the original image, 
i.e., $\bm{\hat{X}} \approx \bm{X}$.

\textbf{Auto-regressive generation}.
For image generation, the quantized feature map $\bm{\hat{Q}}$ can be modeled as a sequence of discrete tokens,
denoted as a sequence of discrete token indices $\bm{I} \in \{0, \dots, K-1\}^{h \times w}$.
Each token roughly corresponds to an image patch of the size $\frac{H}{h} \times \frac{W}{w}$. 
Thus, the prediction of a token sequence is equivalent to synthesizing an image. 
In practice, we refer to uni-directional Transformer\cite{vaswani2017attention} to predict the image sequence autoregressively as follows:
\begin{equation}
    P(\bm{I}_{\leq i} | \bm{T}) = \prod_{j}^{i}P(\bm{I}_j | \bm{I}_{<j}, \bm{T}),
\end{equation}
where $\bm{T}$ is the text tokens of the caption paired with image $\bm{X}$. 

To introduce positional information of the two modalities in Transformer, we learn two sets of positional embeddings. One is axial positional embeddings \cite{ho2019axial} for the visual sequence from a spatial grid. The other is sequence embeddings as BERT \cite{devlin2018bert} for text sequence. 

\subsubsection{Training with Language and Vision Guidance}

\textbf{Main task}.
One consequent training idea is masked sequence modeling by optimizing the loss for the paired text and image tokens. However, unlike most existing vision-and-language models \cite{tan2019lxmert,ni2021m3p,zhou2021uc2} taking detected regions as an image sequence, our model accepts patch sequence, which will be an inexact alignment with text. Moreover, fine-grained correspondences of image patches and attribute tokens are difficult to be aligned. For example, aligning ``a red crown'' and ``a red belly'' within the detected bird needs to precisely recognize not only the color but also the body parts. To avoid noisy training signals, we do not adopt masked sequence modeling.
Instead, our auto-regressive task minimizes cross-entropy losses for the reconstruction of text and image tokens, respectively~\cite{dalle}, 
\begin{equation}\label{l_txt}
    \mathcal{L}_{txt} = - \mathbb{E}_{\bm{T_i}}\log P(\bm{T}_i|\bm{T}_{<i}),
\end{equation}
\begin{equation}
    \mathcal{L}_{img} = - \mathbb{E}_{\bm{I_i}}\log P(\bm{I}_i|\bm{I}_{<i}, \bm{T}).
\end{equation}

\noindent\textbf{Language guidance.}
The transformer model determines the basic image tokens at the top level, and the autoencoder model holds the convolutional decoder complementing the texture in detail at the bottom level. Training these two models separately implies splitting the generation stream stiffly. To this end, we propose a semantic loss for the token prediction such that the model not only considers the downstream decoding but also improves the ability to capture the relation between text and image.

The CLIP \cite{patashnik2021styleclip} is a vision-and-language representation learning model, trained with 400 million image-text pairs, and has shown excellent visual-semantic alignment capability by achieving superb performance on the task of zero-shot image classification. It is optimized by a symmetric cross-entropy loss over the cosine similarities of a batch of image and text embeddings. 
We leverage CLIP to guide our token prediction, through
\begin{equation}
    \mathcal{L}_{CLIP} = 1 - D(G(\bm{\hat{I}}), \bm{T}),
\end{equation}
where $D$ is the cosine similarity between the CLIP embeddings of its two arguments. 
We adopt the straight-through estimator~\cite{bengio2013estimating} for the gradient back-propagation.

\noindent\textbf{Vision guidance.}
With the text descriptions, our model can replace an entity with other specific entities. For only editing the appearance of an entity, we need to provide the model with the prior information on the original entity's shape. Specifically, we convert the image to grayscale and append the quantized grayscale image tokens $\bm{V} \in \{0, \dots, K-1\}^{h \times w}$ to the text sequence as another condition for the tokens to be manipulated. The grayscale image token sequence $\bm{V}$ shares the positional embeddings with the image token sequence $\bm{I}$, for the same modality and spatial positions. For the identities of vision guidance and input text token sequence, we append two special separation tokens [BOV] and [BOT] to the beginning of them respectively. We apply the cross entropy loss on the vision guidance tokens as well,
\begin{equation}
    \mathcal{L}_{gray} = - \mathbb{E}_{\bm{V_i}}\log P(\bm{V}_i|\bm{V}_{<i}).
\end{equation}
We randomly select 50\% samples to train with vision guidance. The total loss to train the transformer is a combination of the four losses, which can be divided into two parts, including auto-regressive and semantic losses as
\begin{equation}
    \mathcal{L}_{ar} = \lambda_{1}\mathcal{L}_{img} + \lambda_{2}\mathcal{L}_{gray} + \lambda_{3}\mathcal{L}_{txt},
\end{equation}
\begin{equation}
    \mathcal{L}_{total} = \mathcal{L}_{ar} + \lambda_{4}\mathcal{L}_{CLIP},
\end{equation}
where $\lambda_1, \lambda_2, \lambda_3$ and $\lambda_4$ are the balancing coefficients.

\subsubsection{Inference with Entity Guidance}
\label{sec:semantic_alignment_module}

We design a semantic alignment module to locate the image patches to be manipulated by input text automatically in the inference phase. 
The semantic alignment module is a two-step module, (1) to find the tokens of every entity and (2) to select the text-relevant entities to be manipulated, where each step is based on a strong existing model.

In the first step, we refer to entity segmentation \cite{entity_segmentation} to recognize each entity on the original image $\bm{X}$, as Fig.~\ref{fig:framework} shows. The segmentation is implemented on the original image size, and we use the bilinear interpolation to resize the binary mask map of each entity to the same size of latent feature map $\bm{Q}$. The pixels whose values are larger than 0 represent that the tokens at the same position belong to the entity. In our preliminary experiments, we compare the bilinear interpolation with max-pooling for finding the entity tokens. The max-pooling dilates the tokens for the bilinear interpolation, however, due to the stack of convolutions in the first stage, the receptive field of the tokens by max-pooling is beyond the entity area and overlaps with other entities. Thus, we use bilinear interpolation to map the segment mask and token mask for a more precise alignment.

In the second step, we set a text prompt word to select the relevant entities. We leverage the FILIP \cite{yao2021filip}, a CLIP-style model optimized by token level similarity, to calculate the similarities between image token and text token. For example, as Fig.~\ref{fig:framework} shows, we set ``bird'' as a prompt word to search the bird entities in the image, and then we average the similarities between tokens of each entity and the prompt word ``bird''. The entities whose similarities are higher than $\theta$ are text-related entities. 

\subsection{SeMani-Diff}
In this section, we analyze several limitations of SeMani-Trans and propose new models to solve these issues.

\textbf{From uni-direction to multi-directions}.
In SeMani-Trans, we train the transformer model to autoregressively generate image tokens in the uni-direction.
However, this is sub-optimal in image manipulation tasks.
Specifically, when the entity-relevant region is in the later part of the sequence, the model will well-generate the tokens with the help of an informative sub-sequence.
But when the entity-relevant region is in the former part of the sequence, the generation model cannot utilize the information of the later unmasked tokens.
In this scenario, the manipulation capacity is limited.

To this end, we propose to use the multi-direction generation process to fully utilize the information of unmasked regions.
Specifically, we adopt the recently fast-developed denoising diffusion probabilistic models (DDPMs).
Unlikely autoregressive models, DDPMs directly encode the information of the entire image.

\textbf{From local semantic similarity to global}.
In SeMani-Trans, we adopt FILIP to calculate the similarity of each visual token to the prompt word and then average them as the similarity between the entity and the prompt word.
However, the token-level feature extracts more local information, and a simple average may not well-extract the global semantics of the entity.
The original CLIP model is trained with global semantics, but it cannot well-extract the feature of an entity. 
This is due to that the shape of the entity has various types, but the CLIP is trained on the image of a square shape.
If we directly use CLIP to extract features of entity images with masks, the resulting embedding is sub-optimal.
Thus, a fine-tuning of CLIP on the masked image is required for better similarity calculation.

With the above two improvements, we now can design a more powerful framework for the entity-level text-guided image manipulation task, dubbed as SeMani-Diff.
SeMani-Diff takes as inputs an original image, a prompt of the entity, and a target text description.
We first segment the image into several entities and then use the fine-tuned CLIP model to calculate the similarity between the entities and the prompt.
The most similar entity will be identified as the target entity to manipulate and forming the masked input image as the visual condition for the generation model.
Then, conditioned on the target text description, we adopt the diffusion models to manipulate the image.

Compared with SeMani-Trans, SeMani-Diff enjoys better semantic alignment thanks to the globally extracted visual features.
Further, the DDPMs enjoy superior generation capacity compared with the autoregressive prediction pipeline due to the better utilization of unmasked regions of the image.
Note that the overall ideology is shared between SeMani-Trans and SeMani-Diff.
We argue that the pipeline of first locating the entity-relevant region via the semantic alignment module and then implementing manipulation via local editing is crucial for eL-TGIM.
The techniques we use for each module are of course not limited to specific architectures.
In the following, we introduce the architectures and training of SeMani-Diff in detail.

\subsubsection{Architectures}
Formally, DDPMs construct two random processes to connect the distribution of real image $\bm{X}_0\sim p(\bm{X}_0)$ with diagonal Gaussian $\bm{X}_T\sim\mathcal{N}(0,\sigma\mathcal{I})$.
In the forward/diffusion process from $\bm{X}_0$ to $\bm{X}_T$, DDPMs gradually add random Gaussian noises, forming a Markov chain as
\begin{equation}
q\left(\bm{X}_t \mid \bm{X}_{t-1}\right)\coloneqq\mathcal{N}\left(\bm{X}_t ; \sqrt{\alpha_t} \bm{X}_{t-1},\left(1-\alpha_t\right) \mathcal{I}\right).
\end{equation}
When the noise added in each step is small enough, and the process runs a long time enough (with a large $T$).
The final state $\bm{X}_T$ can be well-approximated by $\mathcal{N}(0,\sigma\mathcal{I})$ and the posterior distribution $p(\bm{X}_{t-1} \mid \bm{X}_t)$
can also be approximated by a Gaussian distribution.
DDPMs adopt the neural network to learn the posterior via
\begin{equation}
p_\theta\left(\bm{X}_{t-1} \mid \bm{X}_t\right)\coloneqq\mathcal{N}\left(\mu_\theta\left(\bm{X}_t\right), \Sigma_\theta\left(\bm{X}_t\right)\right).
\end{equation}
With this denoising step, we can generate a real image from the random Gaussian noise iteratively.
In each step, unlike autoregressive predictions, DDPMs directly estimate the less-noised image $\bm{X}_{t-1}$ simultaneously.
In this manner, it can well utilize the information of the overall image.

We leverage the latent diffusion models~\cite{rombach2022high} (LDM) that balance the high-resolution image generation and computation cost by introducing DDPMs in the latent space.
Similar to SeMani-Trans, LDM also adopts an auto-encoder with a different target of compressing the image instead of getting discrete token indices.
Thus the dimension of the resulting latent variables is $3$ instead of $1024$ in transformers.
Then a time-conditional UNet~\cite{ronneberger2015u} is performed on the latent variables to learn denoising.

To incorporate the masked image and the target text description as the conditions for generation, we concatenate the masked image with the noise as the input to the diffusion model and adopt cross-attention layer~\cite{vaswani2017attention} to inject text description. 
Then the posterior now becomes:
\begin{equation}
\begin{aligned}
&p_\theta\left(\bm{X}_{t-1} \mid \bm{X}_t,\bm{T},\bm{M}_{\bm{e}}\right) \\  
\coloneqq&\mathcal{N}\left(\mu_\theta\left(\bm{X}_t,\bm{T},\bm{M}_{\bm{e}}\right), \Sigma_\theta\left(\bm{X}_t,\bm{T},\bm{M}_{\bm{e}}\right)\right).
\end{aligned}
\end{equation}

\subsubsection{Training with Masked Image and Language Guidance}
The training of DDPMs can be formulated as first randomly sampling an image from the training set and randomly generating a Gaussian noise by randomly selecting a noise level $t$.
As we can derive $\mu_{\theta},\Sigma_{\theta}$ from the noise $\varepsilon$~\cite{ho2020denoising},
we can directly train the network to learn the error estimation task.
Specifically, conditioned on the masked image and text description, the network can be trained to estimate the noise added to the image via the L2 loss function as 
\begin{equation}
\mathcal{L}\coloneqq\mathbb{E}_{t \sim[1, T], \bm{X}_0 \sim q\left(\bm{X}_0\right), \varepsilon \sim \mathcal{N}(0, \mathcal{I})}\left[\left\|\varepsilon-\varepsilon_\theta\left(\bm{X}_t, \bm{T},\bm{M}_{\bm{e}},t\right)\right\|^2\right] .
\label{eq:ddpm-loss}
\end{equation}

\textbf{Classifier-free guidance}. 
Ho and Salimans~\cite{ho2021classifier} introduce classifier-free
guidance for better generation of diffusion models.
This technique is also utilized in our DDPMs.
Specifically, when training with Eq.~\eqref{eq:ddpm-loss}, we randomly replace the text guidance and mask guidance with empty guidance $\phi$, performing the unconditional generation.
When inference, the final estimation of $\varepsilon$ is 
\begin{equation}
\hat{\varepsilon}=\varepsilon_{\theta}(\bm{X}_t\mid\phi) + s\cdot (\varepsilon_{\theta}(\bm{X}_t\mid\bm{T},\bm{M}_{\bm{e}})-\varepsilon_{\theta}(\bm{X}_t\mid\phi)),
\end{equation}
where $s$ controls the scale for conditioning.

\subsubsection{Inference with Global Entity Guidance}
We strike the balance between local and global semantic similarities via OVSeg~\cite{liang2022open}.
Specifically, we first utilize a segmentation model to distinguish entities in the image.
Then for each entity, we pad the other regions with 0 values, forming the same shape of the squared image as the input image.
Then this entity image with masks is input to the CLIP model to extract more global visual features.
However, the original CLIP model is trained on natural images without masks.
Thus a fine-tuning stage is needed for CLIP to adapt to images with masks.

To fine-tune the CLIP model, OVSeg collects masked image and entity name pairs from the existing image-caption dataset.
Then the pre-trained CLIP model is fine-tuned on this paired dataset to adapt to the masked images.
Specifically, instead of fine-tuning the CLIP model, OVSeg introduces a mask prompt $\bm{P}$ such that the input of the CLIP model becomes $\bm{X}\odot \bm{M} + \bm{P}\odot (1-\bm{M})$ instead of $\bm{X}\odot \bm{M}$, then $\bm{P}$ is learned to better fit the CLIP model and the parameters of CLIP are frozen.
This mask prompt tuning~\cite{jia2022visual} technique is beneficial as the training of mask prompt is easier and preferable when we only have a small training set, compared with fine-tuning the CLIP model.

With this newly fine-tuned CLIP model, the visual feature of the entity can be better extracted, and thus the semantic similarity is more global and semantic than the original local averaged similarity.
The other parts of the semantic alignment module are the same as in SeMani-Trans.

\section{Experiments}
\subsection{Experimental Setup}
\textbf{Competitors}.
As there are only a few methods that can be directly adapted to the task of eL-TGIM, we compare the most related methods that can be used for eL-TGIM without significant modifications.
Specifically, we compare SeMani-Trans and SeMani-Diff against  ManiGAN~\cite{li2020manigan}, Lightweight-GAN~\cite{li2020lightweight}, and Blended LDM~\cite{avrahami2022blended_latent}.
Note that Blended LDM works for the user-provided mask.
In our experiments, we use the mask generated by our semantic alignment model as the input of Blended LDM.
Thus the comparison of Blended LDM and our methods is mostly on the image manipulation phase.
Results of competitors are reproduced using the code/model released by the authors.

\textbf{Datasets}.
Following common practice, we conduct experiments on three public datasets, including  CUB~\cite{cub200}, Oxford~\cite{oxford102}, and the more complicated COCO~\cite{mscoco} datasets. 
The CUB and Oxford are two datasets about birds and flowers respectively. 
CUB contains 8855 training images and 2933 testing images while Oxford has 7034 training images and 1155 testing images, in which each image has 10 captions.
There are at least 80 categories of objects with different shape structures and appearances on COCO images, forming the 80k training images and 40k testing images.
Thus, COCO is a more complicated dataset than CUB and Oxford, not only in the understanding of the correspondence between the image and text but also in image manipulation on the entity level.
We preprocess these datasets as in~\cite{zhang2017stackgan,xu2018attngan}.

\begin{figure*}
\centering
\includegraphics[width=0.9\textwidth]{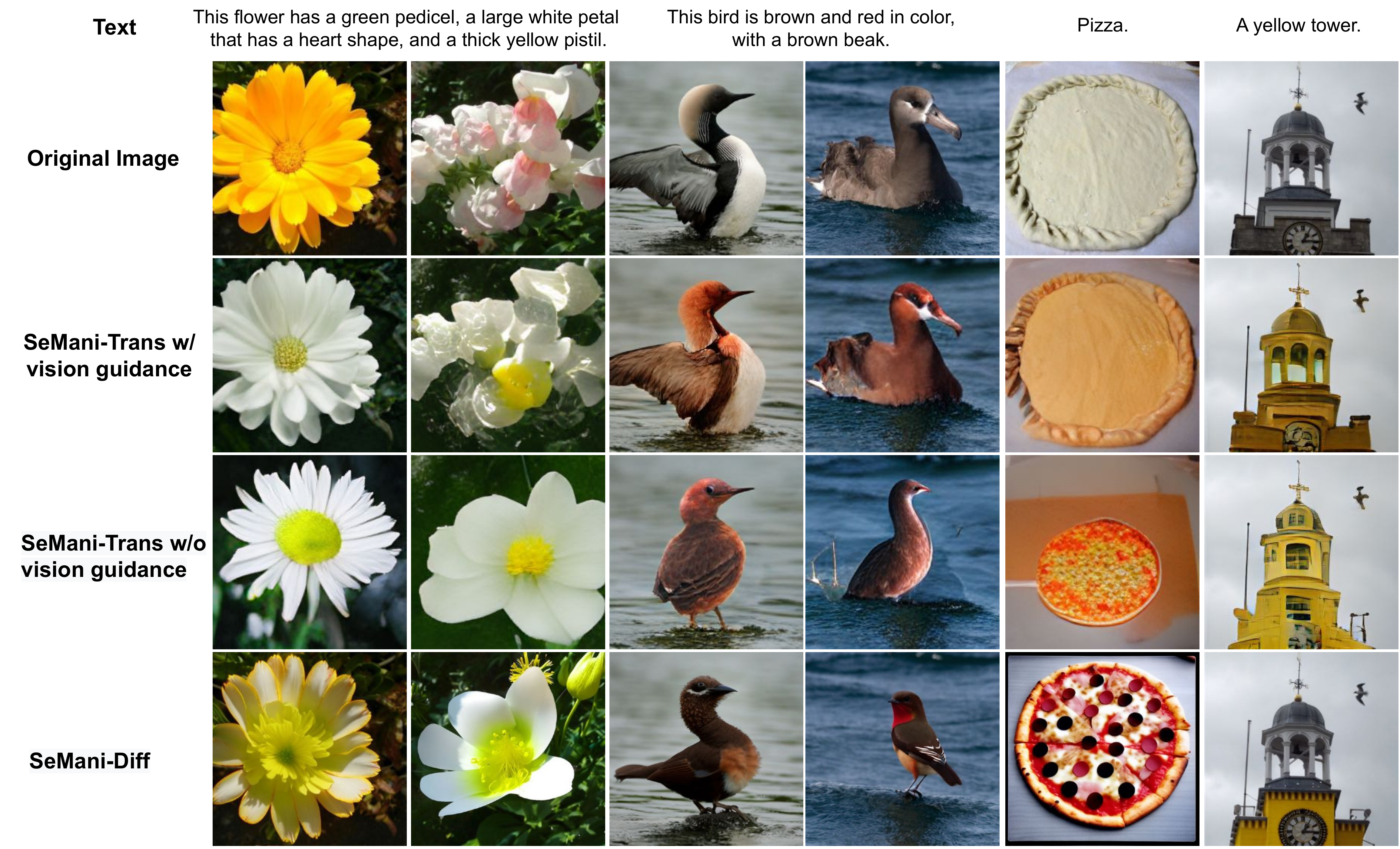}
\vspace{-0.2cm}
\caption{Manipulation results with (w/) and without (w/o) vision guidance. 
SeMani-Trans preserves the structure of the entity when w/ vision guidance. 
SeMani-Trans and SeMani-Diff flexibly perform manipulation according to the text w/o vision guidance. 
The prompt word for the ``Pizza.'' is ``dough''.
}
\label{fig:main_results}
\end{figure*}

\textbf{Quantitative metrics}.
To evaluate the quality of manipulated images, we use the Inception Score (IS) \cite{salimans2016IS} as the quantitative evaluation metric. To evaluate the visual-semantic alignment between the text descriptions and manipulated images, we calculate the cosine similarity between their embeddings extracted with CLIP text/image encoders, called CLIP-score. 
Besides, we conduct an image-to-text retrieval experiment and report Recall@N for quantitative comparison. In the image-to-text retrieval, for each manipulated image, the text candidates consist of the input text, which serves as the positive sample, and 99 randomly sampled descriptions as negative samples. Such 100 text candidates are sorted in descending order according to their cosine similarity with the manipulated image. Recall@N calculates the percentage of images, whose positive sample occurs within the top-N candidates.
As we use the ViT-B/32 CLIP model during training SeMani-Trans, for a fair comparison, we refer to the ResNet50 CLIP model to compute the CLIP-score. 
Additionally, following \cite{tagan}, to compare the quality of the content preservation, we compute the L2 reconstruction error by forwarding images with positive texts. 

The higher the IS, the higher quality of the manipulated images. Higher CLIP-score and R@N indicate better visual-semantic alignment between the input texts and the manipulated images. The lower the L2 error, the higher content preservation quality.

\begin{figure*}
\centering
\vspace{-0.5cm}
\includegraphics[width=0.9\textwidth]{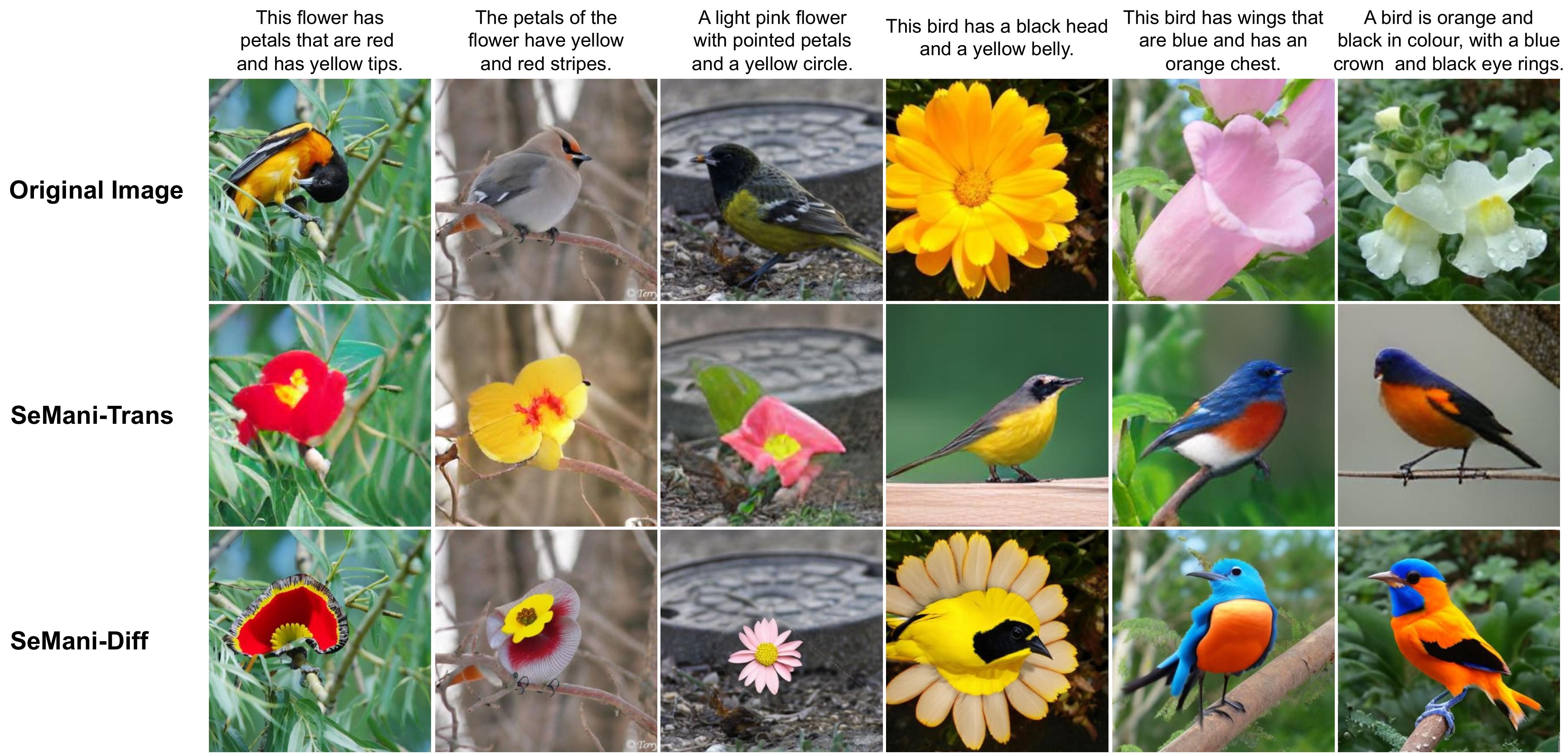}
\vspace{-0.2cm}
\caption{Manipulation results from bird to flower and flower to bird with our proposed SeMani.\label{fig:birdflowerexchange}}

\end{figure*}

\begin{table*}
\centering
\vspace{-0.5cm}
\caption{
Quantitative comparison between ManiGAN~\cite{li2020manigan}, Lightweight-GAN~\cite{li2020lightweight}, Blended LDM~\cite{avrahami2022blended_latent}, and our SeMani-Trans and SeMani-Diff. IS: Inception Score. CLIP-score: averaged cosine similarity with CLIP embeddings. R@10: recall within the top 10 candidates. L2-error: L2 reconstruction error. Higher IS, CLIP-score, and R@10 indicate better performance, while lower L2-error is better.
The best results are in bold.
\label{table:main-results}}
\vspace{-0.3cm}
\resizebox{\textwidth}{!}{
\begin{tabular}{lcccccccccccc}
\toprule
\multirow{2}{*}{Model} & \multicolumn{4}{c}{CUB}                    & \multicolumn{4}{c}{Oxford}                 & \multicolumn{4}{c}{COCO} \tabularnewline                   
& IS        & CLIP-score & R@10   & L2-error & IS        & CLIP-score & R@10   & L2-error & IS         & CLIP-score & R@10   & L2-error \tabularnewline   
\midrule
ManiGAN                & 4.19 \small{$\pm$}  0.04 & 21.30     & 10.49 & 0.05    & 4.37 \small{$\pm$}  0.11 & 21.59     & 14.21 & 0.02    & 22.65 \small{$\pm$}  0.40 & 11.91     & 14.50 & 0.03    \tabularnewline  
Lightweight-GAN        & 4.66 \small{$\pm$}  0.06 & 18.88     & 10.00 & 0.13    &    4.35 \small{$\pm$} 0.09       & 17.55    & 11.58 & 0.12    & 24.80 \small{$\pm$}  0.94 & \textbf{13.65}     & 14.49 & 0.03    \tabularnewline
Blended LDM & \textbf{5.95 \small{$\pm$} 0.08} & 23.34 & 38.17 & 0.02 & 4.21 \small{$\pm$} 0.09 & 22.11 & 21.19 & 0.01 & 27.84 \small{$\pm$} 0.63 & 13.02 & \textbf{28.64} & 0.01\tabularnewline
\midrule
SeMani-Trans                   & 5.02 \small{$\pm$}  0.11 & 23.56     & 34.82 & \textbf{0.01}    & \textbf{4.50 \small{$\pm$}  0.06} & \textbf{23.34}   & \textbf{36.49} & 0.03   & 21.45 \small{$\pm$}  0.41 & 13.10    & 21.32 & 0.02   \tabularnewline
SeMani-Diff & 5.13 \small{$\pm$} 0.07 & \textbf{24.03} & \textbf{45.51} & 0.02 & 4.30 \small{$\pm$} 0.08 & 22.05 & 17.37 & \textbf{0.01} & \textbf{32.98 \small{$\pm$} 0.61} & 12.28 & 24.73 & \textbf{0.01}\tabularnewline
\bottomrule
\end{tabular}
}
\end{table*}

\begin{figure*}
\centering
\includegraphics[width=1\textwidth]{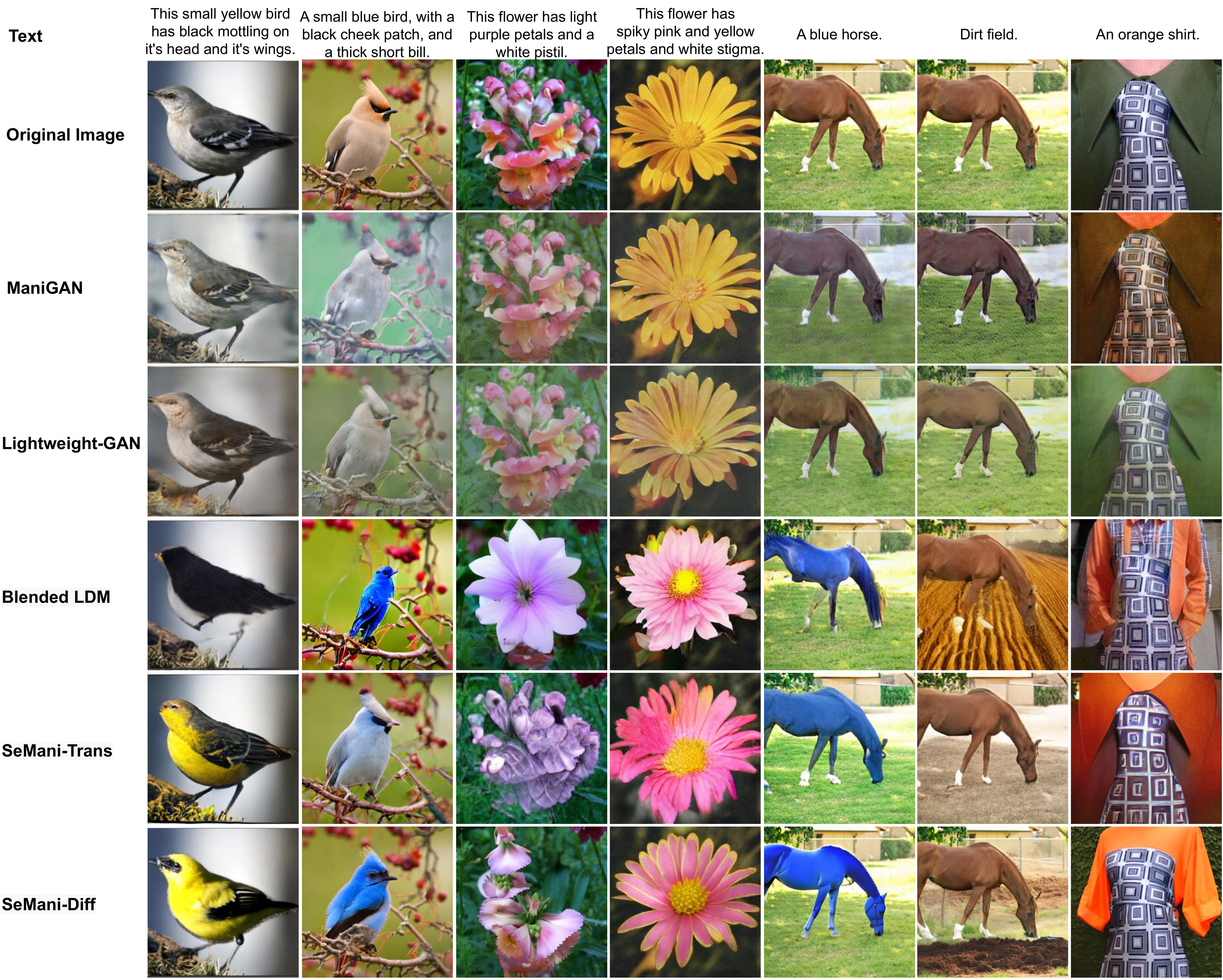}
\caption{Qualitative comparison of different methods on the CUB, Oxford, and COCO datasets. 
SeMani-Trans uses vision guidance to manipulate the images.
Note that Blended LDM uses the entity mask provided by SeMani-Diff as it requires user-provided masks.}
\label{fig:comparewithgan}
\end{figure*}

\textbf{Hyper-parameters of SeMani-Trans}.
The model at the first stage inherits from the VQGAN \cite{vqgan} pre-trained on ImageNet, where the codebook size is 1024, the image size is $256 \times 256$, and the latent feature map size is $16 \times 16$. In the second stage, our transformer has 24 layers, 8 heads with 64 dimensionalities for each head. We replace the traditional Feed-Forward Network (FFN) with a GEGLU \cite{shazeer2020glu} variant, which adds a Gated Linear Units (GLU) \cite{dauphin2017language} with GELU \cite{hendrycks2016gaussian} activation to the first hidden layer of FFN. 
We use Byte-Pair Encoding \cite{sennrich2015neural} to tokenize the text, with vocabulary size 49408. We limit the text length to 128 and learn a padding token for each position as DALL$\cdot$E.
Our transformer has 152M parameters, a little larger than BERT-Base 110M. The hyper-parameters of autoregressive loss $\lambda_1, \lambda_2, \lambda_3$ are set to $7/9, 1/9, 1/9$ and $\lambda_4$ of language guidance loss is $5$ for all the datasets. 
The CLIP model for the semantic loss is ViT-B/32. 
For the semantic alignment module, we use the entity segmentation model based on Swin-L-W7 and the FILIP-large\cite{yao2021filip} model for similarity computation.  The similarity threshold $\theta$ is 0.163.
For a good initialization of the transformer, we pre-train our transformer on CC12M \cite{changpinyo2021cc12m} without language and vision guidance. We use AdamW optimizer with $\beta_1=0.9, \beta_2=0.96$ to train 12 epochs with batch size 112. The learning rate linearly ramps up to $6 \times e^{-4}$ for the first 5k iterations and is halved whenever training loss does not decrease for 50000 iterations. With the same optimizer, we fine-tune our model on the three datasets with the same two steps. The first step fine-tunes the model without vision guidance. The second step adds the vision guidance into training with 50\% samples. Each step lasts 500 epochs with batch size 96 and the learning rate linearly ramps up to $5 \times e^{-4}$ for the first 1k iterations and is halved when training loss does not improve for 10 epochs.

\textbf{Hyper-parameters of SeMani-Diff}.
In SeMani-Diff, we adopt the text-guided inpainting model of latent diffusion models~\cite{rombach2022high} trained on LAION-5B~\cite{schuhmannlaion} dataset as our generative model, as it is more suitable for entity manipulation task.
The diffusion model has 866M parameters.
We use the segmentation model and fine-tuned CLIP model in OVSeg~\cite{liang2022open}.
In the image manipulation phase, we adopt the DDIM~\cite{song2020denoising} generation process with 50 steps with a classifier-free guidance scale $s=9$.

As discussed in Section~\ref{sec:semantic_alignment_module}, we set a word prompt for the entity to be manipulated in the inference phase. Particularly, CUB and Oxford have specific category images, where we set ``bird" and ``flower" as the prompt word respectively. COCO contains various category entities, and we randomly set a prompt word based on the original caption of each image to ensure the entity exists in the image and randomly select a caption of other images as the target text description. Almost all the prompt words of COCO are the nouns of their text in the following experiments and we will clarify the prompt words for special examples.

\subsection{Main Results}
In this section, we first qualitatively verify the manipulation ability of our model to edit or change the entity on the CUB, Oxford, and COCO datasets. As Fig.~\ref{fig:main_results} shows, SeMani-Trans can manipulate the images with the same object structure providing the vision guidance, i.e. the grayscale image, as prior shape information. Without the constraint of the vision guidance, our model generates a different entity corresponding to the text description in place of the original entity. SeMani-Trans merges the generation ability to the manipulation without any user manual mask but only the guidance of input text, where most existing models fail. 
SeMani-Diff enjoys better background preservation compared with SeMani-Trans (for example the first column).
SeMani-Diff also enjoys high-fidelity manipulation consistent with the provided text description.
As SeMani-Diff does not have a vision guidance module, it can not preserve the shape of the entity.

We also conduct an experiment of SeMani-Trans that trained on the mixture of datasets CUB and Oxford to verify a wider manipulation than on the same category. 
As shown in Fig.~\ref{fig:birdflowerexchange}, SeMani-Trans generates reasonably manipulated entities which are corresponding to the text and fit the background, in both bird-to-flower and flower-to-bird settings. For example, 
in the third column from the left, SeMani-Trans not only generates a flower consistent with the description but also complements the upper left corner of the manipulated flower with a leaf, 
which shows that SeMani-Trans also learns a combination of the object information and the background. 
As SeMani-Diff is trained on the large-scale dataset, it also has the capacity of generating a new category.

\subsection{Comparison with the State of the Art}
\textbf{Quantitative results}.
Table \ref{table:main-results} shows the quantitative comparison of our method against previous methods, including 
ManiGAN \cite{li2020manigan}, Lightweight-GAN \cite{li2020lightweight}, and Blended LDM~\cite{avrahami2022blended_latent}. 
Note that
Blended LDM uses the entity mask provided by SeMani-Diff as it requires user-provided masks.
\textbf{(1)} Compare SeMani-Trans with GAN competitors:
On CUB and Oxford datasets, our SeMani-Trans achieves better results than other models on almost all metrics, except for the L2-error on the Oxford dataset, where SeMani-Trans is competitive with ManiGAN. 
It demonstrates that our method can generate high-quality manipulated images (IS), which are consistent with the text descriptions (CLIP-score and R@10) 
, and preserve the content of original images (L2-error).
For the more complicated dataset, COCO, SeMani-Trans outperforms the ManiGAN and Lightweight-GAN on the R@10 and L2-error and achieves a competitive CLIP-score. The IS of our method is competitive with ManiGAN and Lightweight-GAN.  
However, as Fig.~\ref{fig:comparewithgan} shows, 
within many text-guided manipulation cases, ManiGAN and Lightweight-GAN both change the images slightly, more like applying a filter, while SeMani-Trans conducts manipulation according to the text. Typically, the former one is easier to generate high-quality images than the latter
and this is why their IS are a bit higher than our method. 
\textbf{(2)}
On CUB, SeMani-Diff enjoys a much better CLIP-score and R@10 compared with Blended LDM, while on Oxford the performance diverges.
In COCO, SeMani-Diff has a much better IS score than Blended LDM, which can be confirmed in Fig.~\ref{fig:comparewithgan} that Blended LDM introduces unrealistic edges around the manipulated entity (for example the first and second columns) while SeMani-Diff won't.
\textbf{(3)}
In summary, SeMani-Trans and SeMani-Diff show superior or comparable eL-TGIM capacity compared with other methods.

\textbf{Qualitative results}.
As Fig.~\ref{fig:comparewithgan} shows,  
compared with the original images, ManiGAN directs the images toward the semantics of text closer than Lightweight-GAN but changes the background style further from the original as well. Lightweight-GAN preserves the irrelevant contents better than ManiGAN while failing in transforming the text-relevant regions according to the descriptions. 
Blended LDM can generate entities that are consistent with text description, but will introduce unrealistic edges around the entities.
SeMani-Trans outperforms them both on background preservation and foreground manipulation. As the 
second and third columns from the right of the Fig.~\ref{fig:comparewithgan} show, our SeMani-Trans can manipulate the horse and the field respectively on one image, while the baseline methods only change the whole image style with the text.
SeMani-Diff enjoys a higher quality of manipulated results.

\begin{table}
\centering
\caption{The average rank of user study between ManiGAN \cite{li2020manigan}, Lightweight-GAN \cite{li2020lightweight}, Blended LDM~\cite{avrahami2022blended_latent}, and our SeMani-Trans and SeMani-Diff.
Cons., Pre., and Fid. are the abbreviations for consistency, preservation, and fidelity, respectively.
Note that
Blended LDM uses the entity mask provided by SeMani-Diff as it requires user-provided masks.
The lower rank indicates better performance.
\label{table:user-study}}
\begin{tabular}{lccc}
\toprule
Model & Con. ($\downarrow$) & Pre. ($\downarrow$) & Fid. ($\downarrow$)\tabularnewline                   
\midrule
ManiGAN &3.27 & 3.76 & 2.60\tabularnewline  
Lightweight-GAN &4.49 & 4.43 & 4.28\tabularnewline
Blended LDM &2.79 & 2.40 & 3.16\tabularnewline
\midrule
SeMani-Trans &2.68 & 2.93 & 3.03\tabularnewline
SeMani-Diff &\textbf{1.78} & \textbf{1.48} & \textbf{1.93}\tabularnewline
\bottomrule
\end{tabular}
\end{table}

\textbf{User study}.
We conduct a user study experiment to obtain the subjective evaluation of humans.
Specifically, we randomly select 20 testing cases and perform eL-TGIM using our algorithms and competitors.
Then for each case, we design three different perspectives of quality assessment:
(1) \emph{Consistency} which reflects the first requirement of eL-TGIM, to check whether the manipulated image is consistent with the text description;
(2) \emph{Preservation} which reflects the second requirement of eL-TGIM, to ensure that the manipulated image will preserve the text-irrelevant regions;
(3) \emph{Fidelity} which generalizes the third requirement of eL-TGIM, to judge whether the manipulated image is indistinguishable from real pictures.
Then the 20 groups are divided into 5 questionnaires, each of which contains $4\times3=12$ ranking questions to rank the manipulated images (the lower the better) according to the above three perspectives.
Participants are invited to answer one or several questionnaires, and 45 valid questionnaires were recovered. 
The screenshot of the questionnaires can be founded in the appendix, and one of the questionnaires can be found via \url{https://www.wjx.cn/vm/P2IJ5RF.aspx} for convenience.

Results of averaged ranking results are shown in Table~\ref{table:user-study}.
While SeMani-Trans show comparable results compared with diffusion models, our SeMani-Diff achieves significantly superior performance compared with other methods.
This further validates the effectiveness of SeMani and its compatibility with diffusion models.

\begin{figure}
\centering
\includegraphics[width=\linewidth]{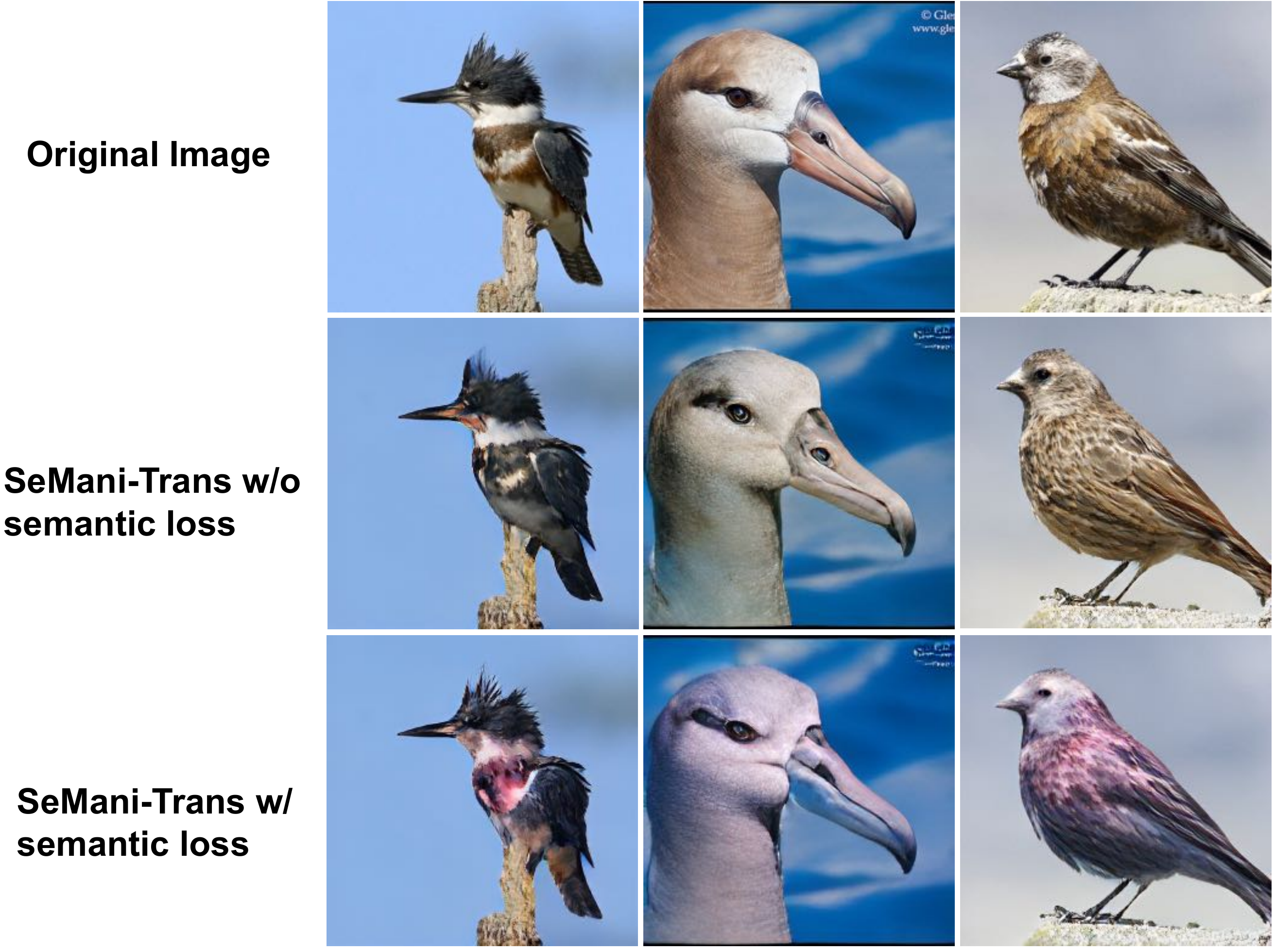}
\vspace{-0.5cm}
\caption{SeMani-Trans w/ and w/o semantic loss on CUB dataset. The text is ``This particular bird has a belly that is purple and gray.''.}
\vspace{-0.3cm}
\label{fig:semantic}
\end{figure}

\subsection{Ablation of SeMani}
\textbf{Semantic loss in SeMani-Trans}.
A comparison of SeMani-Trans trained with and without the semantic loss is shown in Fig.~\ref{fig:semantic}. 
The model trained with the semantic loss manipulates the bird as gray and purple, while the model trained without the semantic loss neglects the purple. It implies that semantic loss helps the model capture the relation between image and text. 

\begin{figure}
\centering
\includegraphics[width=\linewidth]{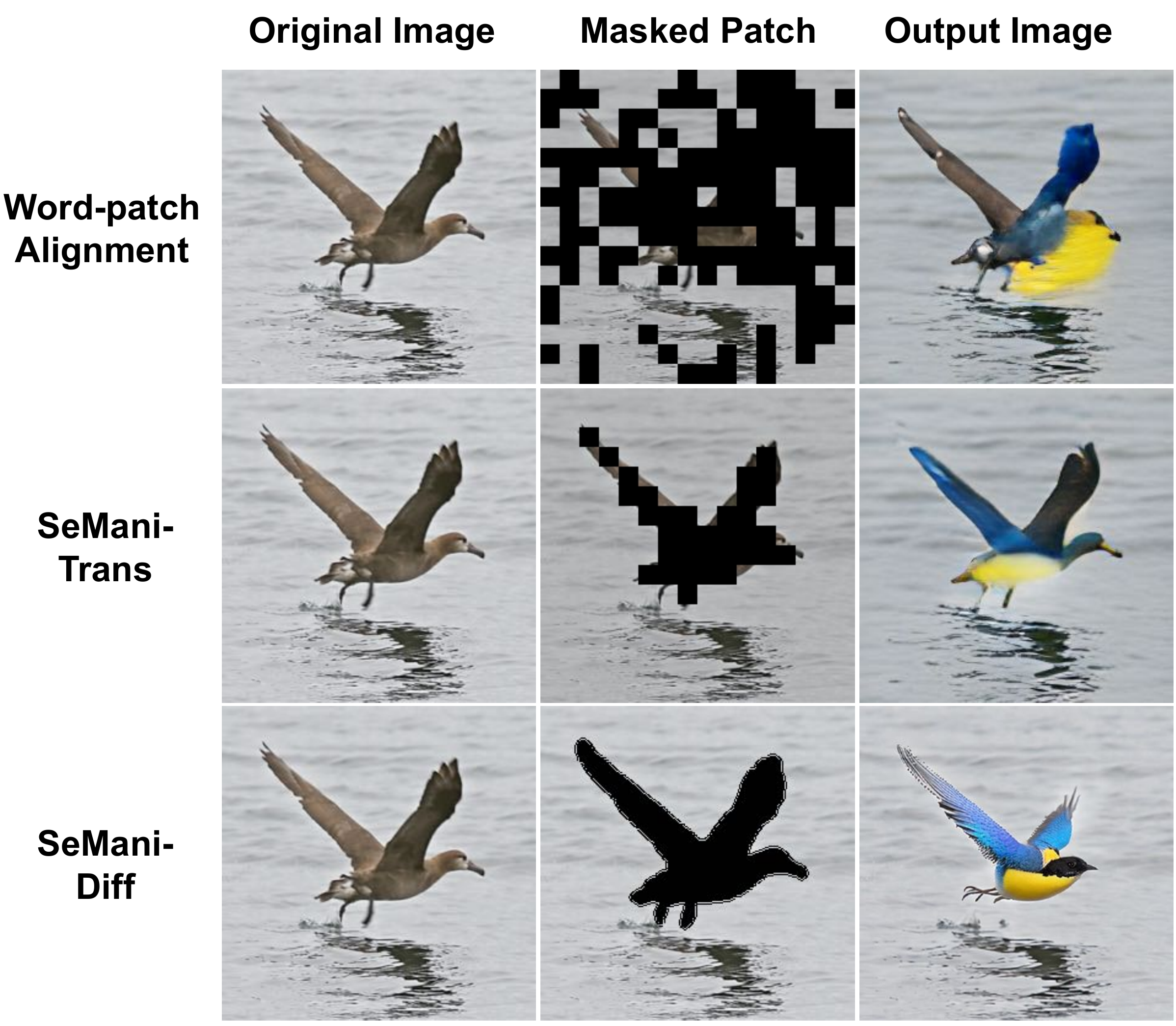}
\vspace{-0.5cm}
\caption{Qualitative comparison of our methods with our semantic alignment mechanism and word-to-patch alignment on the CUB dataset. The text is ``This bird has wings that are blue and has a yellow belly.''.}
\vspace{-0.3cm}
\label{fig:align}
\end{figure}

\textbf{Semantic alignment}.
Word-patch alignment is a technique to align a pair of text and image tokens used in many multi-modality transformer methods \cite{radford2021clip, yao2021filip}.  The word-patch alignment begins from the word tokens to sort the patch tokens, which takes the image patches separately and neglects the information of the entity tokens as a whole during the alignment. Thus the selected image tokens may well scatter within or around an entity area. Manipulating these scattered tokens gets a messy image, where the foreground stays while the background changes. A comparison between word-patch alignment and our semantic alignment method is shown in Fig.~\ref{fig:align}.
The two methods share the same similarity threshold $\theta$ for sorting the image tokens.
As Fig.~\ref{fig:align} shows, our semantic alignment selects the image patches corresponding to the bird precisely, while the word-patch alignment misses some patches corresponding to the bird and selects a few patches which belong to the background. With the inaccurate patches selected by word-patch alignment, only the right-wing turn to blue, and the yellow leaks out. Although the color of the two manipulated images both match the description, the qualitative result by the semantic alignment is better, resulting from more precise edited locations.
Besides, the SeMani-Diff enjoys a more precise alignment in the pixel space, indicating the superiority of the fine-grained control of manipulating region.

\begin{figure}
\centering
\includegraphics[width=\linewidth]{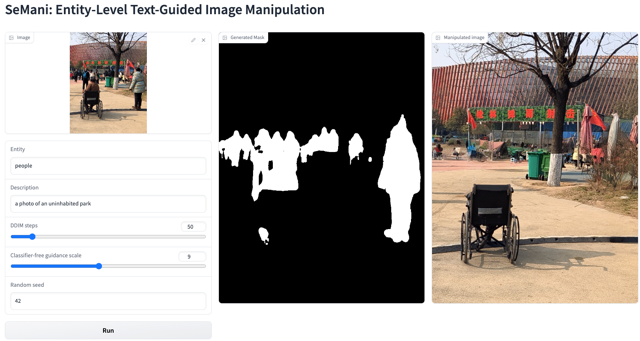}
\vspace{-0.5cm}
\caption{Illustration of our interface for eL-TGIM.
The left part is the input area, where users can input an \emph{image}, an \emph{entity} to manipulate, a text \emph{description} to guide the manipulation, and several hyper-parameters to modify.
Then our model will output the generated mask of the entity in the middle part, and the manipulated image in the right part.
}
\label{fig:interface}
\end{figure}

\subsection{Interface of SeMani}
We design an interface of SeMani for users to perform eL-TGIM with little effort.
Our interface is constructed with Gradio~\cite{abid2019gradio}.
As shown in Fig.~\ref{fig:interface},
the left part of the interface is the input area, where users can upload an original \emph{image}, an \emph{entity} to manipulate, and a text \emph{description} to guide the manipulation.
We also provide several hyper-parameters of the generation model to modify to better match the needs of users.
Then our model will output the generated mask of the entity in the middle part, and the manipulated image in the right part.
A demo video of using the interface to generate results in Fig.~\ref{fig:teaser-diffusion} is in the supplementary material.

\section{Conclusion}

For the first time, this paper studies a new task -- entity-level text-guided image manipulation. 
To tackle this task, we propose a novel framework -- SeMani for the semantic manipulation of eL-TGIM.
Two variants of SeMani from discrete and continuous viewing of images are proposed, respectively.
SeMani-Trans proposes token-wise semantic alignment and manipulation, while SeMani-Diff directly performs semantic alignment and image manipulation at the pixel-level.
Experiments on CUB, Oxford, and COCO validate the superiority of SeMani.

\bibliographystyle{IEEEtran}
\bibliography{mani.bib}

\begin{figure*}
\centering
\includegraphics[width=\linewidth]{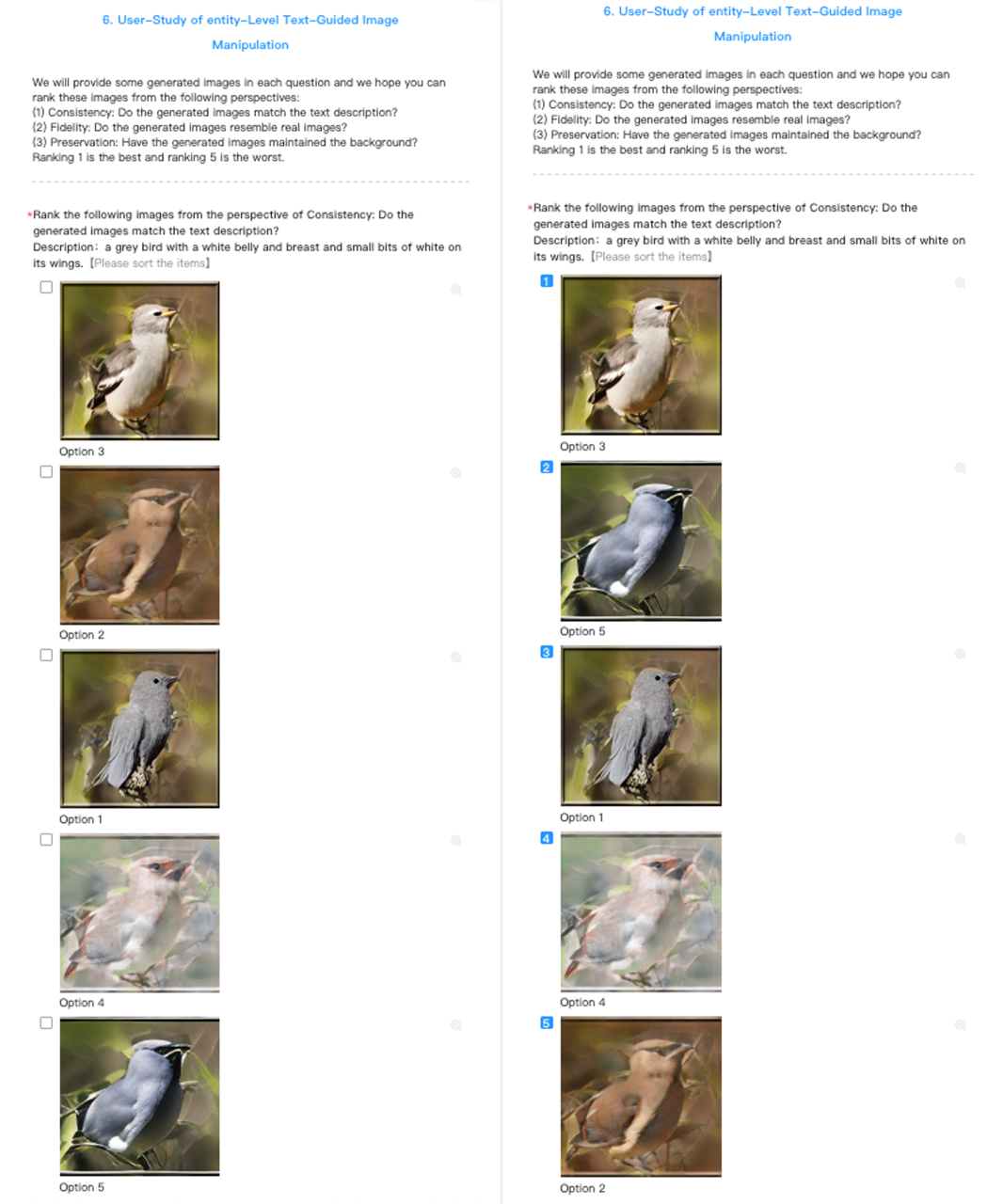}
\caption{Screenshot of our questionnaire used in our user study experiment.
The left part is the interface that participants will see, and the right part is the interface after the ranking is finished by the participants. One of the questionnaires can be found via \url{https://www.wjx.cn/vm/P2IJ5RF.aspx} for convenience.
}
\label{fig:questionnaire}
\end{figure*}

\begin{figure*}
\centering
\includegraphics[width=0.8\linewidth]{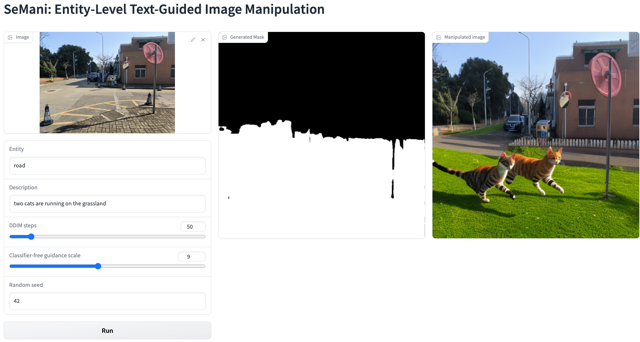}
\includegraphics[width=0.8\linewidth]{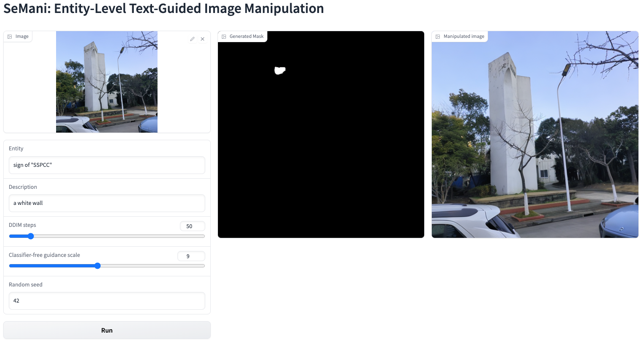}
\includegraphics[width=0.8\linewidth]{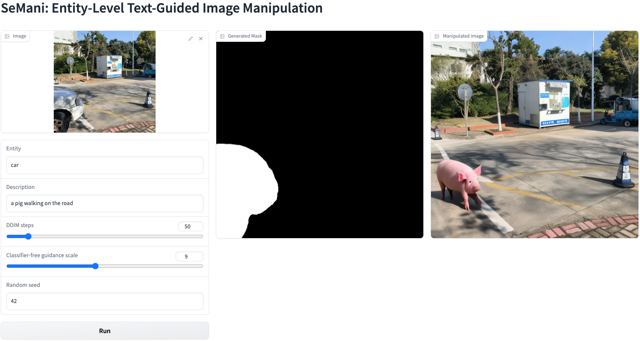}
\caption{More examples of using our interface to perform eL-TGIM.
}
\label{fig:interface-more}
\end{figure*}

\end{document}